\pdfoutput=1

\documentclass[11pt]{article}
\usepackage{CJKutf8}
\usepackage{EMNLP2022}

\usepackage{times}
\usepackage{latexsym}

\usepackage[T1]{fontenc}

\usepackage[utf8]{inputenc}

\usepackage{microtype}

\usepackage{inconsolata}

\usepackage[ruled,vlined]{algorithm2e}
\usepackage{graphicx}
\usepackage{booktabs}
\usepackage{amsmath}
\usepackage{bbm}
\usepackage{multirow}
\usepackage{stfloats}
\usepackage{caption}
\usepackage{subcaption}
\usepackage{longtable}

\title{Evaluating the Efficacy of Length-Controllable Machine Translation}

\author{
  Hao Cheng\textsuperscript{\rm{1}}\thanks{{} {} Work done during the internship at Huawei Noah's Ark Lab.}, Meng Zhang\textsuperscript{\rm{2}}, Weixuan Wang\textsuperscript{\rm{3}}, Liangyou Li\textsuperscript{\rm{2}},  {\bf  Qun Liu\textsuperscript{\rm{2}}}, {\bf Zhihua Zhang\textsuperscript{\rm{4}}} \\
  \textsuperscript{1} Academy for Advanced Interdisciplinary Studies, Peking University \\
  \textsuperscript{2} Huawei Noah's Ark Lab \\
  \textsuperscript{3} Artificial Intelligence Application Research Center, Huawei Technologies \\
  \textsuperscript{4} School of Mathematical Sciences, Peking University \\
  \texttt{hao.cheng@pku.edu.cn} \\
  \texttt{\{zhangmeng92, wangweixuan2, liliangyou, qun.liu\}@huawei.com} \\
  \texttt{zhzhang@math.pku.edu.cn}
}

\begin{document}
\maketitle
\begin{abstract}
Length-controllable machine translation is a type of constrained translation. It aims to contain the original meaning as much as possible while controlling the length of the translation. We can use automatic summarization or machine translation evaluation metrics for length-controllable machine translation, but this is not necessarily suitable and accurate. This work is the first attempt to evaluate the automatic metrics for length-controllable machine translation tasks systematically. We conduct a rigorous human evaluation on two translation directions and evaluate 18 summarization or translation evaluation metrics. We find that BLEURT and COMET have the highest correlation with human evaluation and are most suitable as evaluation metrics for length-controllable machine translation.
\end{abstract}

\section{Introduction}
In the past decade, neural machine translation has benefited from the rapid development of deep learning, significantly improving translation quality~\citep{sutskever2014sequence, bahdanau2014neural, vaswani2017attention}.
Typically, researchers measure the quality of translations by adequacy and fluency, and rarely consider other factors.
For some special translation scenarios, the translation length is constrained, and the length of the generated translation is also used as one of the measurement criteria.
For neural machine translation, it is very challenging to generate a translation of a specified length while maintaining the quality of the translation. The length-controllable translation task has also gradually attracted the attention of the community.
Table~\ref{tab:example} shows an example of length-controllable translation. 

Length-controllable translation has many practical application scenarios. 
Obtaining the gist of a foreign message is a common requirement. The current imperfect translation model can somewhat fulfill the purpose, but it is not designed with the gisting purpose in mind. Length-controllable machine translation can cater to varying degrees of gisting needs. Length-controllable translation can also reduce cognitive load in time-critical scenarios, such as video subtitles and simultaneous translation. Additionally, it can implement rate control in machine interpretation.

\begin{CJK*}{UTF8}{gbsn}
\begin{table}[t]
\begin{tabular}{l|l}
\toprule
Source   & \begin{tabular}[c]{@{}l@{}}Tomorrow is the individual perfor-\\mances that come forward, and \\ that is a different thing to do.\end{tabular} \\\midrule
Reference &\begin{tabular}[c]{@{}l@{}}明天是一场自告奋勇的个人表\\演，与平时有 些不同。\end{tabular}\\ \midrule
80\%                      & \begin{tabular}[c]{@{}l@{}}明天是个人表演，这是一件不\\同的事情。\end{tabular}                                                                                                                                         \\ 
50\%                      & 明天是个人表演，不一样。                                                                                                                                               \\ \bottomrule
\end{tabular}
\caption{An example of length-controllable machine translation, where the specified length is 80\%/50\% of the length of the reference sentence.}
\label{tab:example}
\end{table}
\end{CJK*}

Existing works on length-controllable machine translation are focused on the model~\citep{lakew2019controlling,niehues2020machine,yang2020predicting}. They primarily use BLEU~\citep{papineni2002bleu} or its variant as the evaluation metric\footnote{\citet{niehues2020machine} also used RUSE~\citep{shimanaka-etal-2018-ruse}, but we are unable to obtain the model to evaluate the metric.}. 
Some similar tasks, such as length-controllable summarization~\citep{kikuchi2016controlling, fan2017controllable} and cross-lingual summarization~\citep{bai2021unifying} use ROUGE~\citep{lin2004rouge} as the evaluation metric.
It is simple to directly use evaluation metrics of other tasks on the length-controllable machine translation, but this is not necessarily suitable and accurate for the following reasons. 
In the process of proposing these automatic metrics, the authors measured the correlation with human ratings based on outdated (for now) systems and datasets~\citep{2022Repairing}. More importantly, these systems are not built for length-controllable machine translation.

In this paper, we systematically analyze the existing evaluation metrics. Firstly, we conduct a rigorous manual evaluation. We use two popular length-controllable machine translation models to obtain translations with different lengths. 
Three professional annotators score each translation. Secondly, we evaluate 18 metrics by measuring correlations with human ratings at system and segment levels. Finally, we evaluate three length-controllable translation systems with the recommended automatic metrics.

Our manual evaluation method is relatively novel and different from the manual evaluation for summarization~\citep{koto2022ffci, koto2021evaluating}. Here are some data. SRC: original text.
HT (human translation): manual translation of the original text. 
HS (human summary): the result of summarizing HT manually, equivalent to manual length-controllable translation. 
HYP (hypothesis): the length-controllable translation by the system. In general, manual evaluation of summarization uses HYP and HS. 
However, length-controllable MT requires output with various possible lengths in practical application.
Preparing HS with various lengths is costly, so we use HYP and HT for manual evaluation, and ask annotators to account for the reduced lengths in their ratings.
There is also similar research~\citep{fabbri2021summeval} that uses HYP and HT for manual evaluation. In  their work, HT is the original document to be summarized. Document-level summarization is very difficult to evaluate. The inter-annotator agreement is very low, even though \citet{fabbri2021summeval} employ professional annotators. 
HT in length-controllable MT is sentence-level, which is more feasible for manual evaluation.

The main contributions of this work can be summarized as follows:
    \begin{itemize}
        \item To the best of our knowledge, our work is the first systematic evaluation of length-controllable machine translation.
        \item We evaluate 18 different metrics and find that BLEURT and COMET have  the highest average correlation with human ratings.
        \item We conduct an evaluation of three length-controllable translation systems with the recommended metrics.
    \end{itemize}

\section{Related Work}
\paragraph{Length-Controllable Machine Translation}\:
\citet{lakew2019controlling} address the problem of controlling the output length in machine translation for the first time. ~\citet{niehues2020machine} proposes some other approaches to length-controllable translation. ~\citet{yang2020predicting} propose length prediction as an auxiliary task and set up a sub-network to obtain the length information from the encoder.
~\citet{li2020explicit} study sentence compression to improve the translation quality.
Automatic dubbing~\citep{federico2020speech, tam2021prosody, lakew2021machine, lakew2022isometric, karakanta202042} is a special kind of length-controllable machine translation application case. 
The translation should match the given length to allow synchronization between the source and the target. 

\paragraph{Length-Controllable Text Generation}\:
There are also other tasks for controlling the length of text generation.
\citet{deng2020length} study the length-controllable image captioning. \citet{kikuchi2016controlling, fan2017controllable, liu2018controlling, saito2020length, 2019How, liu2022learning} study monolingual length-controllable summarization. \citet{bai2021unifying} study cross-lingual summarization with compression rate, which is similar to the length-controllable translation task in form. However, summarization typically has a very low ratio between output and input length. Even for sentence-level summarization, the ratio in the data used by \citet{kikuchi2016controlling} is around 30\%, but applications in length-controllable MT typically have much higher ratio. For example, \citet{niehues2020machine} uses ratios of 80\% and 50\%.

\paragraph{Metrics Evaluation}\:
WMT Metrics Shared Task has been held for many years, and new metrics with high correlation with human ratings are constantly proposed. However, these are all evaluations of unrestricted machine translation.
\citet{fabbri2021summeval} first re-evaluate the summarization evaluation. \citet{koto2022ffci} propose a framework for fine-grained summarization evaluation with faithfulness, focus, coverage and inter-sentential coherence.
\citet{koto2021evaluating} study the summarization metrics across languages.

\section{Evaluation Metrics}
We select a total of 18 different metrics commonly used in machine translation, summarization, and text generation research for evaluation.

\paragraph{BLEU*}
BiLingual Evaluation Understudy (BLEU)~\citep{papineni2002bleu} is one of the most commonly used evaluation metrics in machine translation. It measures the $n$-gram match-based precision between the reference and the hypothesis. 
We follow ~\citet{lakew2019controlling} and compute BLEU* by multiplying the BLEU score by the inverse of the brevity penalty.
BLEU* measures the extent to which shorter translations are subsets of longer references.
We use the SacreBLEU~\citep{post-2018-call} implementation\footnote{https://github.com/mjpost/sacrebleu}.

\paragraph{ROUGE} Recall-Oriented Understudy for Gisting Evaluation (ROUGE)~\citep{lin2004rouge} has become one of the most mainstream evaluation metrics for automatic summarization. 
It counts the number of overlapping units such as $n$-gram between the automatically generated summary and the human-written summary.
We consider nine scores (recall, precision, F1) of three variants: ROUGE1 (unigram), ROUGE2 (bigram), and ROUGEL (longest common subsequence). 
We use the \texttt{google-research} implementation\footnote{https://github.com/google-research/google-research/tree/master/rouge}.

\paragraph{BERTScore}
BERTScore~\citep{zhang2019bertscore} 
computes a similarity score for each token in the candidate sentence with each token in the reference sentence using contextual embeddings.
BERTScore computes precision, recall, and F1 measure for evaluating different language generation tasks. 
Researches~\citep{peters-etal-2018-deep, zhang2019bertscore, DBLP:journals/corr/abs-1904-02954} show that it is important to select a good layer or a good combination of layers to generate contextual embedding for different tasks and models. We consider evaluating all layers of four multilingual models (\texttt{bert-base-multilingual-cased}, \texttt{facebook/bart-large-mnli}, \texttt{roberta-large-\\mnli}, \texttt{microsoft/deberta-xlarge-mnli})
and use the official implementation\footnote{https://github.com/Tiiiger/bert\_score}.

\paragraph{BLEURT}
BiLingual Evaluation Understudy with Representations from Transformers (BLEURT)~\citep{sellam2020bleurt} is a metric trained on ratings data. It is a regression model based on BERT~\citep{devlin-etal-2019-bert} and RemBERT~\citep{2020Rethinking}. We consider the recommended checkpoint: BLEURT-20. 
BLEURT-20 is a pre-trained RemBERT model fine-tuned on ratings from the WMT Metrics Shared Task and synthetic data. 
We use the official implementation\footnote{https://github.com/google-research/bleurt}.

\paragraph{COMET}
Crosslingual Optimized Metric for Evaluation of Translation (COMET)~\citep{rei-etal-2020-comet} is a framework for training highly multilingual and adaptable machine translation evaluation models that can function as metrics. We consider three different models.
\texttt{wmt20-comet-da}: best performing metric from WMT20.
\texttt{wmt21-comet-mqm}: best performing metric from WMT21 MQM benchmark.
\texttt{wmt21-comet-qe-da}: referenceless metric trained on DA's from WMT15 to WMT20.
We use the official implementation\footnote{https://github.com/Unbabel/COMET}.

Except BLEU* is system-level, other metrics are segment-level. Segment-level metrics calculates the average of sentence scores to obtain the system score.

\section{Experimental Setup}
\subsection{Data}
We consider two translation directions, En (English) - Zh (Chinese) and Zh-En. 
For each translation direction, we randomly sample 270 sentences from newstest2019 to construct our annotation data. 
We use two popular length-controllable translation models (Length Embedding and Translate-then-Summarize described in Section \ref{subsec:Model}) to generate translation.
For each sentence, we generate two translations of different lengths: 80\% and 50\% of the reference length (length counted based on subwords). The existing length-controllable translation tasks are all shorter translation scenarios and applications, so our experiments is also for shorter translation.
The total number of annotations is 6480 = 2 languages $\times$ 270 sentences $\times$ 2 models $\times$ 2 lengths $\times$ 3 annotators.
The models are trained on WMT17 En-Zh (20M sentence pairs), validated on newsdev2017, and evaluated on newstest2019\footnote{Note that starting from 2019, the two translation directions are different data and both source-original to avoid the translationese effect.}.

\subsection{Annotation}
We use the Appraise Evaluation Framework\footnote{https://github.com/AppraiseDev/Appraise}~\citep{federmann-2018-appraise} as our annotation platform.
Appraise is an open-source framework for crowd-based annotation tasks, notably for evaluation of machine translation.
We create the segment-level direct assessment task~\citep{2013Continuous}.
For data with a length of 80\% annotators are asked to answer the following question: \emph{How much you agree that: the black text expresses 80\% of the main information in the gray text (paraphrasing is acceptable)}.
Annotators use a slider to score and the rating can be anywhere between 0 and 100. In our instruction to annotators, we emphasized the importance of main information. For example, if the black text expresses 80\% of the gray text information but the lost 20\% is the important information of the gray text, in this case, the the rating should not be 100. Appendix~\ref{app:system} shows an example of the annotation task. Appendix~\ref{app:annotator} provides information of the annotators.

We added some trap samples as one of the means to detect the annotation quality. 
We randomly select a sentence from the dataset and truncate it according to the required length (80\% or 50\% of the original target length), keeping only the preceding words. This truncated sentence and the original target form an annotation pair.
We construct 60 trap samples for each annotator. So there are 720 trap annotations = 2 languages $\times$ 60 trap samples $\times$ 2 lengths $\times$ 3 annotators.

\begin{table}[t]
\centering
\begin{tabular}{c|c|cc}
\toprule
Direction              & Length & All.Ave.  & Cut.Ave. \\ \midrule
\multirow{2}{*}{En-Zh} & 80\% & 74.30 & 59.85    \\
                       & 50\% & 90.63 & 53.81    \\ \midrule
\multirow{2}{*}{Zh-En} & 80\% & 90.57 & 42.76    \\
                       & 50\% & 47.84 & 30.43    \\ \bottomrule
\end{tabular}
\caption{The average annotation time (seconds) for each annotation task.}
\label{tab:mean-times}
\end{table}

\begin{table}[t]
\begin{tabular}{@{}ccccccc@{}}
\toprule
      & \multicolumn{2}{c}{Zero} & \multicolumn{2}{c}{Low} & \multicolumn{2}{c}{High} \\\cmidrule{2-7} 
      & 80\%        & 50\%       & 80\%       & 50\%       & 80\%        & 50\%       \\ \midrule
En-Zh & 180         & 176        & 0          & 2          & 0           & 2          \\
Zh-En & 110         & 140        & 53         & 28         & 17          & 12         \\ \bottomrule
\end{tabular}
\caption{The number of trap samples scored at different levels for each annotation task. Zero means rating $=0$, Low means $0<$ rating $\leq20$, and High means rating $>20$.}
\label{tab:trap-rating}
\end{table}

\subsection{Machine Translation Systems}\label{subsec:Model}
We evaluate three length-controllable translation systems using the recommended automatic metrics.
\paragraph{Target Embedding}
This approach from \citet{niehues2020machine} is the best according to their evaluation. It integrates the length constraint directly into the decoder by incorporating the information of the number of remaining target words at each target position. For both this approach and the length embedding approach, the maximum specifiable length is 100 (larger lengths are treated as 100).

\paragraph{Length Embedding}
This approach is similar to the source embedding method from~\citet{niehues2020machine}, but we prepend an additional length token to the target sentence during training. During inference,  we first force decode the desired length token before normal decoding.

\paragraph{Translate-then-Summarize}
This is a simple cascaded model consisting of a standard translation model and a length-controllable summarization model. 
First, obtain the translation by the translation model, and then modify the translation by the summarization model to control the length.

We use Transformer base as the translation model~\citep{vaswani2017attention} in the three systems. And the summarization model is an unsupervised approach~\citep{schumann-etal-2020-discrete}.

\begin{table}[t]
\centering
\begin{tabular}{cccc}
\toprule
\multirow{2}{*}{Direction} & \multirow{2}{*}{Length} & \multicolumn{2}{c}{Pearson correlation ($r$)} \\ \cmidrule(l){3-4} 
                           &                         & w. Trap              & w.o. Trap               \\ \midrule
\multirow{2}{*}{En-Zh}     & 80\%                    & 0.8607              & 0.7724                \\
                           & 50\%                    & 0.7825              & 0.6478                \\\midrule
\multirow{2}{*}{Zh-En}     & 80\%                    & 0.6474              & 0.2935                \\
                           & 50\%                    & 0.6237              & 0.3079                 \\ \bottomrule
\end{tabular}
\caption{The average of one-vs-rest Pearson correlation ($r$) for each annotation task.}
\label{tab:human-pearson}
\end{table}

\begin{table}[t]
\centering
\begin{tabular}{cccc}
\toprule
\multirow{2}{*}{Direction} & \multirow{2}{*}{Length} & \multicolumn{2}{c}{Krippendorff's $\alpha$} \\ \cmidrule{3-4} 
                           &                         & w. Trap              & w.o. Trap             \\ \midrule
\multirow{2}{*}{En-Zh}     & 80\%                    & 0.8212              & 0.7152               \\
                           & 50\%                    & 0.7256              & 0.5741               \\\midrule
\multirow{2}{*}{Zh-En}     & 80\%                    & 0.5229              & 0.2259              \\
                           & 50\%                    & 0.5459              & 0.2423            \\ \bottomrule
\end{tabular}
\caption{The Krippendorff's $\alpha$ for each annotation task.}
\label{tab:human-krippendorff-alpha}
\end{table}

\begin{table*}[t]
\centering
\begin{tabular}{l|ccccc}
\toprule
                                & \multicolumn{2}{c}{En-Zh} & \multicolumn{2}{c}{Zh-En} & \multirow{2}{*}{Ave.} \\
                                & 80\%          & 50\%          & 80\%          & 50\%          &                       \\\midrule
ROUGE                        &             &             &             &             &                       \\\midrule
\quad ROUGE1-P                      & 0.2643      & 0.2299      & 0.2639      & 0.2705      & 0.2571                \\
\quad ROUGE1-R                      & 0.3266      & 0.2431      & 0.2517      & 0.2560      & 0.2694                \\
\quad ROUGE1-F1                      & 0.3168      & 0.2396      & 0.2599      & 0.2622      & 0.2696                \\
\quad ROUGE2-P                      & 0.2612      & 0.2632     & 0.1445      & 0.2196      & 0.2221                \\
\quad ROUGE2-R                      & 0.2911      & 0.2677      & 0.1507      & 0.2129      & 0.2306                \\
\quad ROUGE2-F1                      & 0.2831      & 0.2666      & 0.1484      & 0.2152      & 0.2283                \\
\quad ROUGEL-R                      & 0.3315      & 0.2344      & 0.1959      & 0.2171      & 0.2447                \\
\quad ROUGEL-P                      & 0.2722      & 0.2224      & 0.2017      & 0.2263      & 0.2307                \\
\quad ROUGEL-F1                      & 0.3188      & 0.2311      & 0.1999      & 0.2209      & 0.2427                \\\midrule
COMET                        &             &             &             &             &                       \\\midrule
\quad \texttt{wmt20-comet-da}               & \textbf{0.5080}      & 0.4405      & 0.3342      & 0.4444      & 0.4318                \\
\quad \texttt{wmt21-comet-mqm}              & 0.4083      & 0.5136      & 0.3446      & 0.4561      & 0.4307                \\
\quad \texttt{wmt21-comet-qe-da}            & 0.3837      & 0.4875      & 0.3169      & 0.4349      & 0.4058                \\\midrule
BERTScore                    &             &             &             &             &                       \\\midrule
\quad \texttt{bert-base-multilingual-cased} & 0.3800      & 0.4278      & 0.4203      & 0.4734      & 0.4254                \\
\quad \texttt{bart-large-mnli}              & 0.2586      & 0.2446      & 0.4560      & 0.5447      & 0.3760                \\
\quad \texttt{roberta-large-mnli}           & 0.2651      & 0.2557      & 0.4160      & 0.5210      & 0.3645                \\
\quad \texttt{deberta-xlarge-mnli}          & 0.2571      & 0.2837      & \textbf{0.4798}      & \textbf{0.5500}      & 0.3927                \\\midrule
BLEU*                        & 0.1904      & 0.2063      & 0.1427      & 0.1826      & 0.1805                \\
BLEURT                       & 0.4837      & \textbf{0.5733}      & 0.2836      & 0.4833      & \textbf{0.4560}                \\
\bottomrule
\end{tabular}
\caption{
The Pearson correlation ($r$) between system-level automatic metrics and human ratings. We omit the institution name in BERTScore models.
BERTScore uses the recommended layer and measurement.
}
\label{tab:system-level}
\end{table*}

\section{Results}
\subsection{Annotation Results}

Table~\ref{tab:mean-times} shows the average annotation time (seconds) for each annotation task.
Cut.Ave. denotes the average time of annotation time less than 600 seconds, and All.Ave denotes that of all annotations.
It is confirmed from the time used that the annotators have performed the tasks carefully.

We show the ratings of trap samples in Table~\ref{tab:trap-rating}.
For En-Zh, most trap samples get a zero rating, and only two get a high rating.
For Zh-En, the number of non-zero ratings is relatively large, but most of them are low ratings. 
We check the annotations with non-zero ratings manually and find that most of them are reasonable, because the trap samples are randomly constructed.

Following standard practice in direct assessment, we compute the z-scores from each annotator and then average them for each annotation task.
We calculate the average of one-vs-rest Pearson correlation ($r$) to evaluate the consistency of manual annotation~\citep{amidei2019agreement, koto2022ffci}. 
We analyze the results with and without trap samples respectively.
As we can see from Table~\ref{tab:human-pearson}, the correlation of 50\% is lower than 80\%. 
On the one hand, the 50\%-length-controllable translation task is more challenging. On the other hand, it is more difficult to do the annotation task and judge whether the translation contains 50\% of the main information.
It can be seen that all correlations of En-Zh are higher, which means that the assessment of En-Zh is more reliable. 
We also show Krippendorff's $\alpha$ in Table~\ref{tab:human-krippendorff-alpha}, which is another measure of consistency among annotators. We can observe the same results as Pearson correlation ($r$).

\begin{table*}[t]
\centering
\begin{tabular}{l|ccccc}
\toprule
                             & \multicolumn{2}{c}{En-Zh} & \multicolumn{2}{c}{Zh-En} & \multirow{2}{*}{Ave.} \\
                             & 80\%          & 50\%          & 80\%          & 50\%          &                       \\\midrule
ROUGE                        &             &             &             &             &                       \\\midrule
\quad ROUGE1-P                      & 0.2973      & 0.2605      & 0.1962      & 0.1328      & 0.2217                \\
\quad ROUGE1-R                      & 0.3157      & 0.2753      & 0.1974      & 0.1252      & 0.2284                \\
\quad ROUGE1-F1                      & 0.3139      & 0.2701      & 0.2014      & 0.1287      & 0.2285                \\
\quad ROUGE2-P                      & 0.2915      & 0.2828      & 0.1920      & 0.1438      & 0.2275                \\
\quad ROUGE2-R                      & 0.2980      & 0.2870      & 0.1930      & 0.1364      & 0.2286                \\
\quad ROUGE2-F1                      & 0.2969      & 0.2850      & 0.1936      & 0.1382      & 0.2284                \\
\quad ROUGEL-P                      & 0.2930      & 0.2696      & 0.2392      & 0.1782      & 0.2450                \\
\quad ROUGEL-R                      & 0.3068      & 0.2794      & 0.2397      & 0.1741      & 0.2500                \\
\quad ROUGEL-F1                      & 0.3030      & 0.2760      & 0.2429      & 0.1770      & 0.2497                \\\midrule
COMET                        &             &             &             &             &                       \\\midrule
\quad \texttt{wmt20-comet-da}               & \textbf{0.4720}      & 0.4264      & 0.3210      & 0.2482      & 0.3669                \\
\quad \texttt{wmt21-comet-mqm}              & 0.3844      & 0.3533      & 0.2660      & 0.2158      & 0.3049                \\
\quad \texttt{wmt21-comet-qe-da}            & 0.4413      & 0.3994      & 0.2460      & 0.1911      & 0.3195                \\\midrule
BERTScore                    &             &             &             &             &                       \\\midrule
\quad \texttt{bert-base-multilingual-cased} & 0.3682      & 0.3220      & 0.2577      & 0.2190      & 0.2917                \\
\quad \texttt{bart-large-mnli}              & 0.2541      & 0.2094      & 0.3121      & 0.2600      & 0.2589                \\
\quad \texttt{roberta-large-mnli}           & 0.2778      & 0.2116      & 0.3024      & 0.2631      & 0.2637                \\
\quad \texttt{deberta-xlarge-mnli}          & 0.2889      & 0.2321      & 0.3084      & 0.2392      & 0.2672                \\\midrule
BLEURT                       & 0.4611      & \textbf{0.4348}      & \textbf{0.3330}      & \textbf{0.2722}      & \textbf{0.3753}                \\
\bottomrule
\end{tabular}
\caption{The Kendall correlations ($\tau$) between segment-level automatic metrics and human ratings. We omit the institution name in BERTScore models.
BERTScore uses the recommended layer and measurement.
}
\label{tab:segment-level}
\end{table*}

\begin{table}[t]
\centering
\begin{tabular}{ccc}
\toprule
{\small{}Model}                        & {\small{}Layer} & {\small{}Type} \\ \midrule
{\small{}\texttt{bert-base-multilingual-cased}} & {\small{}8}     & {\small{}R}    \\
{\small{}\texttt{bart-large-mnli}}              & {\small{}11}    & {\small{}F1}   \\
{\small{}\texttt{roberta-large-mnli}}           & {\small{}13}    & {\small{}F1}   \\
{\small{}\texttt{deberta-xlarge-mnli}}          & {\small{}41}    & {\small{}F1}   \\ \bottomrule
\end{tabular}
\caption{The recommended layer and type of measurement for multilingual BERTScore models.}
\label{tab:bertscore-select}
\end{table}

\subsection{Metrics Evaluation}

\subsubsection{System-Level Evaluation}
Among all the metrics we consider, except BLEU* which directly obtains the score of system level, other metrics (including human) obtain the system-level score by calculating the average of segment-level scores. For system-level evaluation, we use hybrid super sampling~\citep{graham2016achieving}. 
Table~\ref{tab:system-level} shows the Pearson correlation ($r$) between various automatic metrics and human ratings, broken down across language and translation length. For BERTScore, we only show the result of the recommended layer and measurement.

We perform selection for BERTScore by selecting the layer and measurement that produces the highest average correlation over four task.
Table~\ref{tab:bertscore-select} details the recommended layer and measurement for each of the multilingual models.
For the correlation of each layer and measurement of BERTScore models, please refer to Appendix~\ref{app:bertscore}.

As we can see from Table~\ref{tab:system-level}, except for BERTScore, the performance of other metrics is relatively consistent in the four tasks.
BLEU*, the most commonly used in length-controllable translation, performs the worst.
ROUGE1-F1 and ROUGE1-R have the best overall performance among ROUGE variants. 
COMET metrics are better than ROUGE metrics in four tasks, and \texttt{wmt20-comet-da} is overall the best version. 
The performance of BERTScore on En-Zh is relatively poor, but the performance on Zh-En is significantly better than other metrics.
BLEURT performs better than other metrics overall:
Except for the poor performance on the task of 80\%-length-controllable of Zh-En, the performance on other tasks is excellent.

\begin{figure*}[!t]
    \centering
    \begin{subfigure}[c]{0.45\textwidth}
        \centering
        \includegraphics[width=\textwidth]{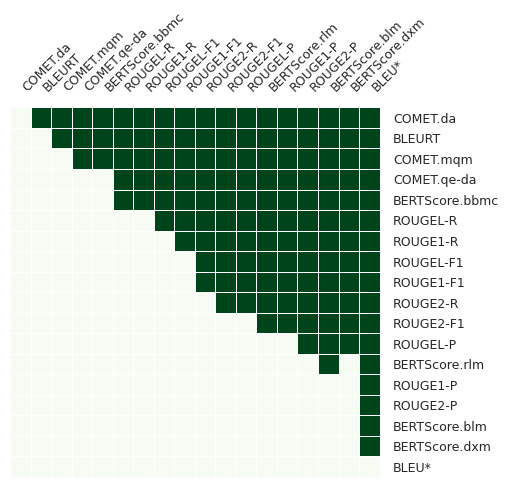}
        \caption{80\%-length, En-Zh}
        \label{fig:sub1}
    \end{subfigure}
    \hfill
    \begin{subfigure}[c]{0.45\textwidth}
        \centering
        \includegraphics[width=\textwidth]{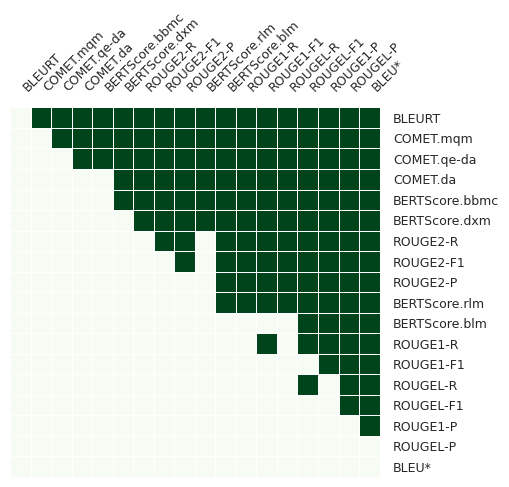}
        \caption{50\%-length, En-Zh}
        \label{fig:sub2}
    \end{subfigure}

    \caption{
    System-level significance tests. Green cells indicate a significant win for the row metric over the column metric, with 95\% confidence intervals of a difference in correlations not containing zero. For BERTScore and COMET, we shortened the names of the models due to page size constraints.}
    \label{fig:system-level}
\end{figure*}

\begin{figure*}[!ht]
    \centering
    \begin{subfigure}[c]{0.45\textwidth}
        \centering
        \includegraphics[width=\textwidth]{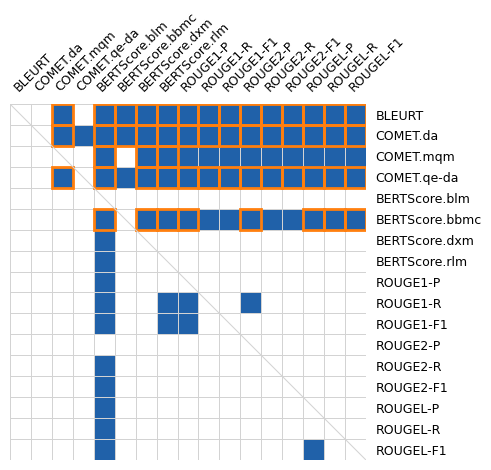}
        \caption{80\%-length, En-Zh}
        \label{fig:seg-sub1}
    \end{subfigure}
    \hfill
    \begin{subfigure}[c]{0.45\textwidth}
        \centering
        \includegraphics[width=\textwidth]{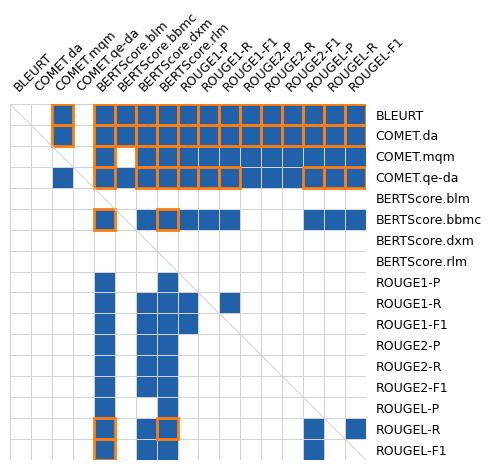}
        \caption{50\%-length, En-Zh}
        \label{fig:seg-sub2}
    \end{subfigure}

    \caption{
Segment-level significance tests. Blue cells indicate the row metric has a higher correlation than the column metric ($p<0.05$).
The orange outline indicates the result remains significant after applying the Bonferroni correction.
}
    \label{fig:segment-level}
\end{figure*}

\subsubsection{Segment-Level Evaluation}
Following~\citep{freitag-etal-2021-results}, we measure correlation using the Kendall $\tau$ statistic for the segment-level evaluation. 

Table~\ref{tab:segment-level} show the Kendall correlations ($\tau$) between automatic metrics and human ratings.
The recommended layer and measurement for BERTScore in Table~\ref{tab:bertscore-select} is also used in segment-level evaluation.
For segment-level evaluation, ROUGEL-F1 and ROUGEL-R have the best overall performance among ROUGE variants.
COMET metrics are still better than ROUGE metrics in the four tasks.  \texttt{wmt20-comet-da} is still the best version. 
BERTScore has mixed performance across models and tasks, but on average it performs worse than COMET and BLEURT.
BLEURT outperforms almost all other metrics on all tasks.

\subsubsection{Significance Testing}

For system-level metrics, we follow~\citet{graham2016achieving} and construct confidence intervals for differences in dependent correlations using the method presented by~\citet{Zou2007Toward}. We use the implementation\footnote{https://github.com/ygraham/MT-metric-confidence-intervals} of~\citet{graham2016achieving} and the result of system level on En-Zh is shown in Figure~\ref{fig:system-level}. Appendix~\ref{app:sign-test} shows the result on Zh-En.

For segment-level metrics, we follow ~\citet{freitag-etal-2021-results} and run PERM-BOTH hypothesis test\footnote{https://github.com/CogComp/stat-analysis-experiments}~\citep{deutsch2021statistical} to find a significant difference between metrics. Figure~\ref{fig:segment-level} shows the result of segment level on En-Zh. Result on Zh-En is also in Appendix~\ref{app:sign-test}.

We can conclude from Table~\ref{tab:system-level} and  Table~\ref{tab:segment-level} that BLEURT and COMET.\texttt{wmt20-comet-da} are the best metrics. For the system-level evaluation, Figure ~\ref{fig:system-level} shows that COMET.\texttt{wmt20-comet-da} significantly wins BLEURT in 80\%-length, but loses in 50\%-length. 
For the segment-level evaluation, as we can see from Figure~\ref{fig:segment-level}, there is no significant difference between the two.

\subsubsection{Metric Recommendation}
Both in system-level and segment-level evaluations, BLEURT and COMET.\texttt{wmt20-comet-da} have the overall best performance.
Although BERTScore has a better performance in the system-level evaluation on Zh-En, we do not recommend using it as a general evaluation metric. In addition, the human rating has a higher correlation on En-Zh, and the results based on En-Zh are more reliable. Therefore, in general, we recommend using BLEURT and COMET.\texttt{wmt20-comet-da} as the metrics of length-controllable machine translation.

\begin{table*}[t]
\centering
\begin{tabular}{l|llll}
\toprule
                    & \multicolumn{2}{c}{En-Zh} & \multicolumn{2}{c}{Zh-En} \\
                    & \multicolumn{1}{c}{80\%}          & \multicolumn{1}{c}{50\%}           & \multicolumn{1}{c}{80\%}           & \multicolumn{1}{c}{50\%}           \\\midrule
BLEURT              &             &             &             &             \\\midrule
Target Embedding    & 0.5882        & 0.4327        & 0.6321        & 0.5104        \\
Length Embedding     &  0.5923$^{\dagger}$       & 0.4444$^{\dagger\dagger}$     & 0.6316       &  0.5132           \\
Translate-then-Summarize &    0.5607     &   0.3804       & 0.5961      &   0.4550        \\\midrule
COMET.\texttt{wmt20-comet-da}               &             &             &             &             \\\midrule
Target Embedding    &  0.2209       & -0.4597       & 0.2398       & -0.4205      \\
Length Embedding    &  0.2415$^{\dagger\dagger}$     &  -0.4012$^{\dagger\dagger}$           &  0.2362   &  -0.4129            \\
Translate-then-Summarize &   0.1509      &     -0.5148     &  0.1115     &   -0.6031    \\ \bottomrule
\end{tabular}
\caption{Evaluation of the three length-controllable machine translation models with BLEURT and COMET.\texttt{wmt20-comet-da}. We use paired bootstrap resampling \citep{koehn-2004-statistical} for significance testing. $\dagger$: $p<0.05$; ${\dagger\dagger}$: $p<0.01$.}
\label{tab:baselines}
\end{table*}

\begin{table*}[t]
\centering
\begin{tabular}{l|cccc}
\toprule
                         & \multicolumn{2}{c}{En-Zh} & \multicolumn{2}{c}{Zh-En} \\
                         & 80\%        & 50\%        & 80\%        & 50\%        \\\midrule
Target Embedding         & 0           & 0           & 7.4e-5      & 3.9e-5              \\
Length Embedding         & 0.019       & 0.021       & 0.019       & 0.018       \\
Translate-then-Summarize & 0.034       & 0.020       & 0.040       & 0.019       \\\bottomrule
\end{tabular}
\caption{Length deviation of the three length-controllable machine translation models.}
\label{tab:length_deviation}
\end{table*}

\subsection{Machine Translation System Evaluation}
We implement the three translation systems described in Section \ref{subsec:Model} and evaluate them with the recommended BLEURT and COMET.\texttt{wmt20-comet-da}. 
The results are shown in Table~\ref{tab:baselines}.
As we can see, Translate-then-Summarize has the worst performance. 
This shows that simply combining translation and summarization models can not achieve length-controllable translation with satisfactory results.
BLEURT and COMET.\texttt{wmt20-comet-da} consistently shows that Length Embedding outperforms Target Embedding on En-Zh.

We use length deviation to measure a model's ability of controlling the length, which is calculated as $|output\_{len} - expect\_{len}|/expect\_{len}$ for each test instance and averaged over the test set.

As we can see from Table~\ref{tab:length_deviation}, Translate-then-Summarize is less effective at controlling length. Target Embedding can precisely control the length of the output due to the design of the model.

\section{Conclusion}
In this work, we evaluate 18 metrics for length-controllable machine translation. 
By comparing their correlation with human ratings, we find that BLEURT and COMET.\texttt{wmt20-comet-da} perform better on four tasks with different controlled lengths and translation directions. 
Therefore, we recommend BLEURT and COMET.\texttt{wmt20-comet-da} as metrics of length-controllable machine translation.


\section*{Limitations}
The experiments in this paper are based on English (En) and Chinese (Zh). Although these are two distant languages, we find BLEURT and COMET to be consistent winners. But we cannot guarantee results on other languages which are far away from En and Zh. Our data are also limited to the news domain. Besides, the annotation consistency of English data is relatively low, indicating the challenge of evaluating natural language generation tasks~\citep{2022Repairing}.



\bibliography{custom}

\begin{thebibliography}{45}
\expandafter\ifx\csname natexlab\endcsname\relax\def\natexlab#1{#1}\fi

\bibitem[{Amidei et~al.(2019)Amidei, Piwek, and Willis}]{amidei2019agreement}
Jacopo Amidei, Paul Piwek, and Alistair Willis. 2019.
\newblock \href {https://doi.org/10.18653/v1/W19-8642} {Agreement is overrated:
  A plea for correlation to assess human evaluation reliability}.
\newblock In \emph{Proceedings of the 12th International Conference on Natural
  Language Generation}, pages 344--354, Tokyo, Japan. Association for
  Computational Linguistics.

\bibitem[{Bahdanau et~al.(2015)Bahdanau, Cho, and Bengio}]{bahdanau2014neural}
Dzmitry Bahdanau, Kyunghyun Cho, and Yoshua Bengio. 2015.
\newblock \href {http://arxiv.org/abs/1409.0473} {Neural machine translation by
  jointly learning to align and translate}.
\newblock In \emph{3rd International Conference on Learning Representations,
  {ICLR} 2015, San Diego, CA, USA, May 7-9, 2015, Conference Track
  Proceedings}.

\bibitem[{Bai et~al.(2021)Bai, Huang, Fan, Gao, Zhu, Zhan, Chi, and
  Chen}]{bai2021unifying}
Yu~Bai, Heyan Huang, Kai Fan, Yang Gao, Yiming Zhu, Jiaao Zhan, Zewen Chi, and
  Boxing Chen. 2021.
\newblock \href {https://arxiv.org/abs/2110.07936} {Unifying cross-lingual
  summarization and machine translation with compression rate}.
\newblock \emph{ArXiv preprint}, abs/2110.07936.

\bibitem[{Chung et~al.(2021)Chung, F{\'{e}}vry, Tsai, Johnson, and
  Ruder}]{2020Rethinking}
Hyung~Won Chung, Thibault F{\'{e}}vry, Henry Tsai, Melvin Johnson, and
  Sebastian Ruder. 2021.
\newblock \href {https://openreview.net/forum?id=xpFFI\_NtgpW} {Rethinking
  embedding coupling in pre-trained language models}.
\newblock In \emph{9th International Conference on Learning Representations,
  {ICLR} 2021, Virtual Event, Austria, May 3-7, 2021}. OpenReview.net.

\bibitem[{Deng et~al.(2020)Deng, Ding, Tan, and Wu}]{deng2020length}
Chaorui Deng, Ning Ding, Mingkui Tan, and Qi~Wu. 2020.
\newblock Length-controllable image captioning.
\newblock In \emph{European Conference on Computer Vision}, pages 712--729.
  Springer.

\bibitem[{Deutsch et~al.(2021)Deutsch, Dror, and Roth}]{deutsch2021statistical}
Daniel Deutsch, Rotem Dror, and Dan Roth. 2021.
\newblock \href {https://doi.org/10.1162/tacl_a_00417} {A statistical analysis
  of summarization evaluation metrics using resampling methods}.
\newblock \emph{Transactions of the Association for Computational Linguistics},
  9:1132--1146.

\bibitem[{Devlin et~al.(2019)Devlin, Chang, Lee, and
  Toutanova}]{devlin-etal-2019-bert}
Jacob Devlin, Ming-Wei Chang, Kenton Lee, and Kristina Toutanova. 2019.
\newblock \href {https://doi.org/10.18653/v1/N19-1423} {{BERT}: Pre-training of
  deep bidirectional transformers for language understanding}.
\newblock In \emph{Proceedings of the 2019 Conference of the North {A}merican
  Chapter of the Association for Computational Linguistics: Human Language
  Technologies, Volume 1 (Long and Short Papers)}, pages 4171--4186,
  Minneapolis, Minnesota. Association for Computational Linguistics.

\bibitem[{Fabbri et~al.(2021)Fabbri, Kry{\'s}ci{\'n}ski, McCann, Xiong, Socher,
  and Radev}]{fabbri2021summeval}
Alexander~R. Fabbri, Wojciech Kry{\'s}ci{\'n}ski, Bryan McCann, Caiming Xiong,
  Richard Socher, and Dragomir Radev. 2021.
\newblock \href {https://doi.org/10.1162/tacl_a_00373} {{S}umm{E}val:
  Re-evaluating summarization evaluation}.
\newblock \emph{Transactions of the Association for Computational Linguistics},
  9:391--409.

\bibitem[{Fan et~al.(2018)Fan, Grangier, and Auli}]{fan2017controllable}
Angela Fan, David Grangier, and Michael Auli. 2018.
\newblock \href {https://doi.org/10.18653/v1/W18-2706} {Controllable
  abstractive summarization}.
\newblock In \emph{Proceedings of the 2nd Workshop on Neural Machine
  Translation and Generation}, pages 45--54, Melbourne, Australia. Association
  for Computational Linguistics.

\bibitem[{Federico et~al.(2020)Federico, Enyedi, Barra-Chicote, Giri, Isik,
  Krishnaswamy, and Sawaf}]{federico2020speech}
Marcello Federico, Robert Enyedi, Roberto Barra-Chicote, Ritwik Giri, Umut
  Isik, Arvindh Krishnaswamy, and Hassan Sawaf. 2020.
\newblock \href {https://doi.org/10.18653/v1/2020.iwslt-1.31} {From
  speech-to-speech translation to automatic dubbing}.
\newblock In \emph{Proceedings of the 17th International Conference on Spoken
  Language Translation}, pages 257--264, Online. Association for Computational
  Linguistics.

\bibitem[{Federmann(2018)}]{federmann-2018-appraise}
Christian Federmann. 2018.
\newblock \href {https://aclanthology.org/C18-2019} {Appraise evaluation
  framework for machine translation}.
\newblock In \emph{Proceedings of the 27th International Conference on
  Computational Linguistics: System Demonstrations}, pages 86--88, Santa Fe,
  New Mexico. Association for Computational Linguistics.

\bibitem[{Freitag et~al.(2021)Freitag, Rei, Mathur, Lo, Stewart, Foster, Lavie,
  and Bojar}]{freitag-etal-2021-results}
Markus Freitag, Ricardo Rei, Nitika Mathur, Chi-kiu Lo, Craig Stewart, George
  Foster, Alon Lavie, and Ond{\v{r}}ej Bojar. 2021.
\newblock \href {https://aclanthology.org/2021.wmt-1.73} {Results of the
  {WMT}21 metrics shared task: Evaluating metrics with expert-based human
  evaluations on {TED} and news domain}.
\newblock In \emph{Proceedings of the Sixth Conference on Machine Translation},
  pages 733--774, Online. Association for Computational Linguistics.

\bibitem[{Gehrmann et~al.(2022)Gehrmann, Clark, and Sellam}]{2022Repairing}
S.~Gehrmann, E.~Clark, and T.~Sellam. 2022.
\newblock \href {https://arxiv.org/abs/2202.06935} {Repairing the cracked
  foundation: A survey of obstacles in evaluation practices for generated
  text}.
\newblock \emph{ArXiv preprint}, abs/2202.06935.

\bibitem[{Graham et~al.(2013)Graham, Baldwin, Moffat, and
  Zobel}]{2013Continuous}
Yvette Graham, Timothy Baldwin, Alistair Moffat, and Justin Zobel. 2013.
\newblock \href {https://aclanthology.org/W13-2305} {Continuous measurement
  scales in human evaluation of machine translation}.
\newblock In \emph{Proceedings of the 7th Linguistic Annotation Workshop and
  Interoperability with Discourse}, pages 33--41, Sofia, Bulgaria. Association
  for Computational Linguistics.

\bibitem[{Graham and Liu(2016)}]{graham2016achieving}
Yvette Graham and Qun Liu. 2016.
\newblock \href {https://doi.org/10.18653/v1/N16-1001} {Achieving accurate
  conclusions in evaluation of automatic machine translation metrics}.
\newblock In \emph{Proceedings of the 2016 Conference of the North {A}merican
  Chapter of the Association for Computational Linguistics: Human Language
  Technologies}, pages 1--10, San Diego, California. Association for
  Computational Linguistics.

\bibitem[{Karakanta et~al.(2020)Karakanta, Negri, and Turchi}]{karakanta202042}
Alina Karakanta, Matteo Negri, and Marco Turchi. 2020.
\newblock \href {https://doi.org/10.18653/v1/2020.iwslt-1.26} {Is 42 the answer
  to everything in subtitling-oriented speech translation?}
\newblock In \emph{Proceedings of the 17th International Conference on Spoken
  Language Translation}, pages 209--219, Online. Association for Computational
  Linguistics.

\bibitem[{Kikuchi et~al.(2016)Kikuchi, Neubig, Sasano, Takamura, and
  Okumura}]{kikuchi2016controlling}
Yuta Kikuchi, Graham Neubig, Ryohei Sasano, Hiroya Takamura, and Manabu
  Okumura. 2016.
\newblock \href {https://doi.org/10.18653/v1/D16-1140} {Controlling output
  length in neural encoder-decoders}.
\newblock In \emph{Proceedings of the 2016 Conference on Empirical Methods in
  Natural Language Processing}, pages 1328--1338, Austin, Texas. Association
  for Computational Linguistics.

\bibitem[{Koehn(2004)}]{koehn-2004-statistical}
Philipp Koehn. 2004.
\newblock \href {https://aclanthology.org/W04-3250} {Statistical significance
  tests for machine translation evaluation}.
\newblock In \emph{Proceedings of the 2004 Conference on Empirical Methods in
  Natural Language Processing}, pages 388--395, Barcelona, Spain. Association
  for Computational Linguistics.

\bibitem[{Koto et~al.(2022)Koto, Baldwin, and Lau}]{koto2022ffci}
Fajri Koto, Timothy Baldwin, and Jey~Han Lau. 2022.
\newblock Ffci: A framework for interpretable automatic evaluation of
  summarization.
\newblock \emph{Journal of Artificial Intelligence Research}, 73:1553--1607.

\bibitem[{Koto et~al.(2021)Koto, Lau, and Baldwin}]{koto2021evaluating}
Fajri Koto, Jey~Han Lau, and Timothy Baldwin. 2021.
\newblock \href {https://doi.org/10.18653/v1/2021.findings-acl.71} {Evaluating
  the efficacy of summarization evaluation across languages}.
\newblock In \emph{Findings of the Association for Computational Linguistics:
  ACL-IJCNLP 2021}, pages 801--812, Online. Association for Computational
  Linguistics.

\bibitem[{Lakew et~al.(2019)Lakew, Di~Gangi, and
  Federico}]{lakew2019controlling}
Surafel~Melaku Lakew, Mattia Di~Gangi, and Marcello Federico. 2019.
\newblock \href {https://aclanthology.org/2019.iwslt-1.31} {Controlling the
  output length of neural machine translation}.
\newblock In \emph{Proceedings of the 16th International Conference on Spoken
  Language Translation}, Hong Kong. Association for Computational Linguistics.

\bibitem[{Lakew et~al.(2021)Lakew, Federico, Wang, Hoang, Virkar,
  Barra-Chicote, and Enyedi}]{lakew2021machine}
Surafel~Melaku Lakew, Marcello Federico, Yue Wang, Cuong Hoang, Yogesh Virkar,
  Roberto Barra-Chicote, and Robert Enyedi. 2021.
\newblock Machine translation verbosity control for automatic dubbing.
\newblock In \emph{ICASSP 2021-2021 IEEE International Conference on Acoustics,
  Speech and Signal Processing (ICASSP)}, pages 7538--7542. IEEE.

\bibitem[{Lakew et~al.(2022)Lakew, Virkar, Mathur, and
  Federico}]{lakew2022isometric}
Surafel~Melaku Lakew, Yogesh Virkar, Prashant Mathur, and Marcello Federico.
  2022.
\newblock Isometric mt: Neural machine translation for automatic dubbing.
\newblock In \emph{ICASSP 2022-2022 IEEE International Conference on Acoustics,
  Speech and Signal Processing (ICASSP)}, pages 6242--6246. IEEE.

\bibitem[{Li et~al.(2020)Li, Wang, Chen, Utiyama, Sumita, Zhang, and
  Zhao}]{li2020explicit}
Zuchao Li, Rui Wang, Kehai Chen, Masao Utiyama, Eiichiro Sumita, Zhuosheng
  Zhang, and Hai Zhao. 2020.
\newblock \href {https://aaai.org/ojs/index.php/AAAI/article/view/6347}
  {Explicit sentence compression for neural machine translation}.
\newblock In \emph{The Thirty-Fourth {AAAI} Conference on Artificial
  Intelligence, {AAAI} 2020, The Thirty-Second Innovative Applications of
  Artificial Intelligence Conference, {IAAI} 2020, The Tenth {AAAI} Symposium
  on Educational Advances in Artificial Intelligence, {EAAI} 2020, New York,
  NY, USA, February 7-12, 2020}, pages 8311--8318. {AAAI} Press.

\bibitem[{Lin(2004)}]{lin2004rouge}
Chin-Yew Lin. 2004.
\newblock \href {https://aclanthology.org/W04-1013} {{ROUGE}: A package for
  automatic evaluation of summaries}.
\newblock In \emph{Text Summarization Branches Out}, pages 74--81, Barcelona,
  Spain. Association for Computational Linguistics.

\bibitem[{Liu et~al.(2022)Liu, Huang, and Mou}]{liu2022learning}
Puyuan Liu, Chenyang Huang, and Lili Mou. 2022.
\newblock \href {https://arxiv.org/abs/2205.14521} {Learning non-autoregressive
  models from search for unsupervised sentence summarization}.
\newblock \emph{ArXiv preprint}, abs/2205.14521.

\bibitem[{Liu et~al.(2018)Liu, Luo, and Zhu}]{liu2018controlling}
Yizhu Liu, Zhiyi Luo, and Kenny Zhu. 2018.
\newblock \href {https://doi.org/10.18653/v1/D18-1444} {Controlling length in
  abstractive summarization using a convolutional neural network}.
\newblock In \emph{Proceedings of the 2018 Conference on Empirical Methods in
  Natural Language Processing}, pages 4110--4119, Brussels, Belgium.
  Association for Computational Linguistics.

\bibitem[{Niehues(2020)}]{niehues2020machine}
Jan Niehues. 2020.
\newblock \href {https://aclanthology.org/2020.amta-research.3} {Machine
  translation with unsupervised length-constraints}.
\newblock In \emph{Proceedings of the 14th Conference of the Association for
  Machine Translation in the Americas (Volume 1: Research Track)}, pages
  21--35, Virtual. Association for Machine Translation in the Americas.

\bibitem[{Papineni et~al.(2002)Papineni, Roukos, Ward, and
  Zhu}]{papineni2002bleu}
Kishore Papineni, Salim Roukos, Todd Ward, and Wei-Jing Zhu. 2002.
\newblock \href {https://doi.org/10.3115/1073083.1073135} {{B}leu: a method for
  automatic evaluation of machine translation}.
\newblock In \emph{Proceedings of the 40th Annual Meeting of the Association
  for Computational Linguistics}, pages 311--318, Philadelphia, Pennsylvania,
  USA. Association for Computational Linguistics.

\bibitem[{Peters et~al.(2018)Peters, Neumann, Iyyer, Gardner, Clark, Lee, and
  Zettlemoyer}]{peters-etal-2018-deep}
Matthew~E. Peters, Mark Neumann, Mohit Iyyer, Matt Gardner, Christopher Clark,
  Kenton Lee, and Luke Zettlemoyer. 2018.
\newblock \href {https://doi.org/10.18653/v1/N18-1202} {Deep contextualized
  word representations}.
\newblock In \emph{Proceedings of the 2018 Conference of the North {A}merican
  Chapter of the Association for Computational Linguistics: Human Language
  Technologies, Volume 1 (Long Papers)}, pages 2227--2237, New Orleans,
  Louisiana. Association for Computational Linguistics.

\bibitem[{Post(2018)}]{post-2018-call}
Matt Post. 2018.
\newblock \href {https://doi.org/10.18653/v1/W18-6319} {A call for clarity in
  reporting {BLEU} scores}.
\newblock In \emph{Proceedings of the Third Conference on Machine Translation:
  Research Papers}, pages 186--191, Brussels, Belgium. Association for
  Computational Linguistics.

\bibitem[{Rei et~al.(2020)Rei, Stewart, Farinha, and
  Lavie}]{rei-etal-2020-comet}
Ricardo Rei, Craig Stewart, Ana~C Farinha, and Alon Lavie. 2020.
\newblock \href {https://doi.org/10.18653/v1/2020.emnlp-main.213} {{COMET}: A
  neural framework for {MT} evaluation}.
\newblock In \emph{Proceedings of the 2020 Conference on Empirical Methods in
  Natural Language Processing (EMNLP)}, pages 2685--2702, Online. Association
  for Computational Linguistics.

\bibitem[{Reimers and Gurevych(2019)}]{DBLP:journals/corr/abs-1904-02954}
Nils Reimers and Iryna Gurevych. 2019.
\newblock \href {https://arxiv.org/abs/1904.02954} {Alternative weighting
  schemes for elmo embeddings}.
\newblock \emph{ArXiv preprint}, abs/1904.02954.

\bibitem[{Saito et~al.(2020)Saito, Nishida, Nishida, Otsuka, Asano, Tomita,
  Shindo, and Matsumoto}]{saito2020length}
Itsumi Saito, Kyosuke Nishida, Kosuke Nishida, Atsushi Otsuka, Hisako Asano,
  Junji Tomita, Hiroyuki Shindo, and Yuji Matsumoto. 2020.
\newblock \href {https://arxiv.org/abs/2001.07331} {Length-controllable
  abstractive summarization by guiding with summary prototype}.
\newblock \emph{ArXiv preprint}, abs/2001.07331.

\bibitem[{Schumann et~al.(2020)Schumann, Mou, Lu, Vechtomova, and
  Markert}]{schumann-etal-2020-discrete}
Raphael Schumann, Lili Mou, Yao Lu, Olga Vechtomova, and Katja Markert. 2020.
\newblock \href {https://doi.org/10.18653/v1/2020.acl-main.452} {Discrete
  optimization for unsupervised sentence summarization with word-level
  extraction}.
\newblock In \emph{Proceedings of the 58th Annual Meeting of the Association
  for Computational Linguistics}, pages 5032--5042, Online. Association for
  Computational Linguistics.

\bibitem[{Sellam et~al.(2020)Sellam, Das, and Parikh}]{sellam2020bleurt}
Thibault Sellam, Dipanjan Das, and Ankur Parikh. 2020.
\newblock \href {https://doi.org/10.18653/v1/2020.acl-main.704} {{BLEURT}:
  Learning robust metrics for text generation}.
\newblock In \emph{Proceedings of the 58th Annual Meeting of the Association
  for Computational Linguistics}, pages 7881--7892, Online. Association for
  Computational Linguistics.

\bibitem[{Sennrich et~al.(2016)Sennrich, Haddow, and
  Birch}]{sennrich2016neural}
Rico Sennrich, Barry Haddow, and Alexandra Birch. 2016.
\newblock \href {https://doi.org/10.18653/v1/P16-1162} {Neural machine
  translation of rare words with subword units}.
\newblock In \emph{Proceedings of the 54th Annual Meeting of the Association
  for Computational Linguistics (Volume 1: Long Papers)}, pages 1715--1725,
  Berlin, Germany. Association for Computational Linguistics.

\bibitem[{Shimanaka et~al.(2018)Shimanaka, Kajiwara, and
  Komachi}]{shimanaka-etal-2018-ruse}
Hiroki Shimanaka, Tomoyuki Kajiwara, and Mamoru Komachi. 2018.
\newblock \href {https://doi.org/10.18653/v1/W18-6456} {{RUSE}: Regressor using
  sentence embeddings for automatic machine translation evaluation}.
\newblock In \emph{Proceedings of the Third Conference on Machine Translation:
  Shared Task Papers}, pages 751--758, Belgium, Brussels. Association for
  Computational Linguistics.

\bibitem[{Sun et~al.(2019)Sun, Shapira, Dagan, and Nenkova}]{2019How}
Simeng Sun, Ori Shapira, Ido Dagan, and Ani Nenkova. 2019.
\newblock \href {https://doi.org/10.18653/v1/W19-2303} {How to compare
  summarizers without target length? pitfalls, solutions and re-examination of
  the neural summarization literature}.
\newblock In \emph{Proceedings of the Workshop on Methods for Optimizing and
  Evaluating Neural Language Generation}, pages 21--29, Minneapolis, Minnesota.
  Association for Computational Linguistics.

\bibitem[{Sutskever et~al.(2014)Sutskever, Vinyals, and
  Le}]{sutskever2014sequence}
Ilya Sutskever, Oriol Vinyals, and Quoc~V. Le. 2014.
\newblock \href
  {https://proceedings.neurips.cc/paper/2014/hash/a14ac55a4f27472c5d894ec1c3c743d2-Abstract.html}
  {Sequence to sequence learning with neural networks}.
\newblock In \emph{Advances in Neural Information Processing Systems 27: Annual
  Conference on Neural Information Processing Systems 2014, December 8-13 2014,
  Montreal, Quebec, Canada}, pages 3104--3112.

\bibitem[{Tam et~al.(2021)Tam, Lakew, Virkar, Mathur, and
  Federico}]{tam2021prosody}
Derek Tam, Surafel~Melaku Lakew, Yogesh Virkar, Prashant Mathur, and Marcello
  Federico. 2021.
\newblock \href {https://arxiv.org/abs/2112.08548} {Prosody-aware neural
  machine translation for dubbing}.
\newblock \emph{ArXiv preprint}, abs/2112.08548.

\bibitem[{Vaswani et~al.(2017)Vaswani, Shazeer, Parmar, Uszkoreit, Jones,
  Gomez, Kaiser, and Polosukhin}]{vaswani2017attention}
Ashish Vaswani, Noam Shazeer, Niki Parmar, Jakob Uszkoreit, Llion Jones,
  Aidan~N. Gomez, Lukasz Kaiser, and Illia Polosukhin. 2017.
\newblock \href
  {https://proceedings.neurips.cc/paper/2017/hash/3f5ee243547dee91fbd053c1c4a845aa-Abstract.html}
  {Attention is all you need}.
\newblock In \emph{Advances in Neural Information Processing Systems 30: Annual
  Conference on Neural Information Processing Systems 2017, December 4-9, 2017,
  Long Beach, CA, {USA}}, pages 5998--6008.

\bibitem[{Yang et~al.(2020)Yang, Gao, Wang, and Ney}]{yang2020predicting}
Zijian Yang, Yingbo Gao, Weiyue Wang, and Hermann Ney. 2020.
\newblock \href {https://aclanthology.org/2020.aacl-main.41} {Predicting and
  using target length in neural machine translation}.
\newblock In \emph{Proceedings of the 1st Conference of the Asia-Pacific
  Chapter of the Association for Computational Linguistics and the 10th
  International Joint Conference on Natural Language Processing}, pages
  389--395, Suzhou, China. Association for Computational Linguistics.

\bibitem[{Zhang et~al.(2020)Zhang, Kishore, Wu, Weinberger, and
  Artzi}]{zhang2019bertscore}
Tianyi Zhang, Varsha Kishore, Felix Wu, Kilian~Q. Weinberger, and Yoav Artzi.
  2020.
\newblock \href {https://openreview.net/forum?id=SkeHuCVFDr} {Bertscore:
  Evaluating text generation with {BERT}}.
\newblock In \emph{8th International Conference on Learning Representations,
  {ICLR} 2020, Addis Ababa, Ethiopia, April 26-30, 2020}. OpenReview.net.

\bibitem[{Zou(2007)}]{Zou2007Toward}
Guang~Yong Zou. 2007.
\newblock Toward using confidence intervals to compare correlations.
\newblock \emph{Psychological Methods}, 12(4):399--413.

\end{thebibliography}
\bibliographystyle{acl_natbib}

\clearpage
\appendix

\section{Data Preprocessing}
\label{app:data}
We segment Chinese by character, and use Moses\footnote{https://github.com/moses-smt/mosesdecoder}
scripts for punctuation normalization
and tokenization of English. The corpora
are deduplicated. Each language is encoded with
byte pair encoding (BPE) \citep{sennrich2016neural}
with 64k merge operations. The BPE codes and
vocabularies are learned on the merged parallel data.
Sentences with more than 128 subwords are removed. Parallel sentences are cleaned with length
ratio 1.5.

\section{Annotation System}
\label{app:system}
\begin{figure*}[t]
\centering
\includegraphics[width=\textwidth]{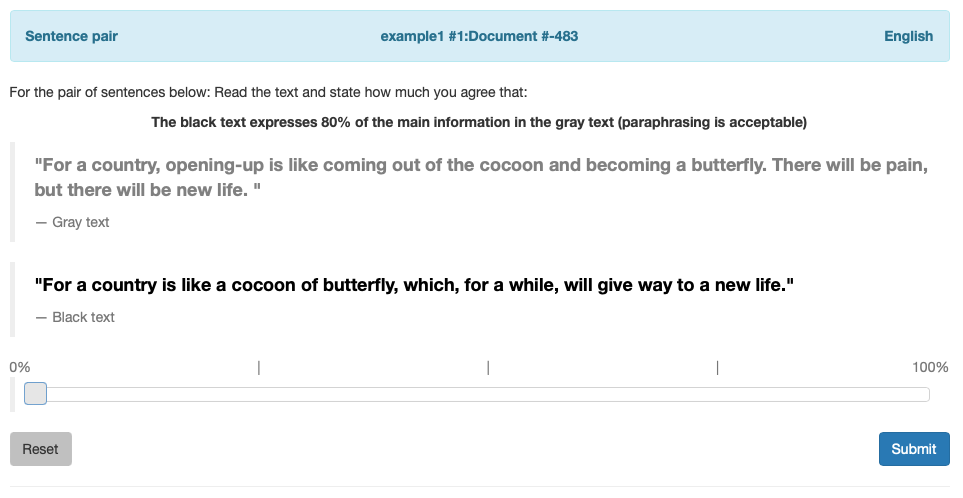}
\caption{An example of 80\%-length-controllable annotation task of Zh-En.}
\label{fig:system}

\end{figure*}

Figure~\ref{fig:system} shows an example of 80\%-length-controllable annotation task of Zh-En.

\section{Annotator Information}
\label{app:annotator}
We have three professional annotators for each language. They are hired by contract and properly paid. All three Chinese annotators are native speakers. Two English annotators are also native speakers; the third English annotator is non-native but has language mastery of English acquired through education and work experience in an English-speaking country.

\section{BERTScore Evaluation}
\label{app:bertscore}
Table~\ref{tab:bertscore.bbmc}, Table~\ref{tab:bertscore.blm}, Table~\ref{tab:bertscore.dxm} and Table~\ref{tab:bertscore.rlm} show the Pearson correlation ($r$) between BERTScore (\texttt{bert-base-multilingual-cased}, \texttt{bart-large-mnli}, \texttt{deberta-xlarge-mnli} and \texttt{roberta-large-mnli}) and human ratings. 
Labels can be formalized as \texttt{layer.\{num\}.\{P/R/F\}}, where \texttt{\{num\}} represents the layer of used features, and \texttt{P}, \texttt{R}, and \texttt{F} represent Precision, Recall, and F1, respectively.

\begin{table*}[]
\centering
\begin{tabular}{lccccc}
\toprule
                             & \multicolumn{2}{c}{Zh-En} & \multicolumn{2}{c}{En-Zh} & \multirow{2}{*}{Ave.} \\
                             & 80\%          & 50\%          & 80\%          & 50\%          &                       \\\midrule
layer.8.R  & 0.4203    & 0.4734    & 0.3800    & 0.4278    & 0.4254 \\
layer.8.F  & 0.4000    & 0.4938    & 0.3605    & 0.4234    & 0.4195 \\
layer.7.R  & 0.4149    & 0.4763    & 0.3727    & 0.4115    & 0.4189 \\
layer.9.R  & 0.4136    & 0.4614    & 0.3850    & 0.4035    & 0.4159 \\
layer.9.F  & 0.3965    & 0.4790    & 0.3650    & 0.4073    & 0.4119 \\
layer.7.F  & 0.3857    & 0.4851    & 0.3576    & 0.4003    & 0.4072 \\
layer.10.R & 0.4070    & 0.4546    & 0.3811    & 0.3822    & 0.4062 \\
layer.12.R & 0.3876    & 0.4758    & 0.3918    & 0.3629    & 0.4045 \\
layer.10.F & 0.3909    & 0.4715    & 0.3621    & 0.3868    & 0.4028 \\
layer.6.R  & 0.4059    & 0.4740    & 0.3567    & 0.3637    & 0.4001 \\
layer.11.R & 0.4039    & 0.4473    & 0.3816    & 0.3572    & 0.3975 \\
layer.12.F & 0.3528    & 0.4811    & 0.3709    & 0.3844    & 0.3973 \\
layer.11.F & 0.3770    & 0.4593    & 0.3641    & 0.3709    & 0.3928 \\
layer.8.P  & 0.3620    & 0.4779    & 0.3251    & 0.3993    & 0.3911 \\
layer.5.R  & 0.4028    & 0.4515    & 0.3605    & 0.3448    & 0.3899 \\
layer.9.P  & 0.3662    & 0.4669    & 0.3305    & 0.3938    & 0.3893 \\
layer.6.F  & 0.3763    & 0.4734    & 0.3414    & 0.3428    & 0.3835 \\
layer.10.P & 0.3623    & 0.4600    & 0.3291    & 0.3740    & 0.3813 \\
layer.4.R  & 0.3934    & 0.4293    & 0.3666    & 0.3339    & 0.3808 \\
layer.3.R  & 0.3855    & 0.4142    & 0.3626    & 0.3244    & 0.3717 \\
layer.7.P  & 0.3309    & 0.4542    & 0.3236    & 0.3699    & 0.3697 \\
layer.11.P & 0.3357    & 0.4434    & 0.3315    & 0.3645    & 0.3688 \\
layer.12.P & 0.2850    & 0.4499    & 0.3380    & 0.3895    & 0.3656 \\
layer.2.R  & 0.3728    & 0.4047    & 0.3630    & 0.3193    & 0.3650 \\
layer.1.R  & 0.3846    & 0.4020    & 0.3598    & 0.3093    & 0.3639 \\
layer.5.F  & 0.3583    & 0.4386    & 0.3422    & 0.3163    & 0.3638 \\
layer.0.R  & 0.3842    & 0.3927    & 0.3555    & 0.2910    & 0.3558 \\
layer.4.F  & 0.3363    & 0.4074    & 0.3484    & 0.3102    & 0.3506 \\
layer.6.P  & 0.3214    & 0.4325    & 0.3058    & 0.3081    & 0.3419 \\
layer.3.F  & 0.3187    & 0.3826    & 0.3470    & 0.3050    & 0.3383 \\
layer.2.F  & 0.3075    & 0.3721    & 0.3484    & 0.3027    & 0.3327 \\
layer.1.F  & 0.3082    & 0.3583    & 0.3447    & 0.2922    & 0.3259 \\
layer.0.F  & 0.3146    & 0.3571    & 0.3390    & 0.2786    & 0.3223 \\
layer.5.P  & 0.2615    & 0.3639    & 0.2978    & 0.2746    & 0.2994 \\
layer.4.P  & 0.1890    & 0.2968    & 0.2996    & 0.2708    & 0.2641 \\
layer.3.P  & 0.1312    & 0.2418    & 0.2986    & 0.2689    & 0.2351 \\
layer.2.P  & 0.1015    & 0.2137    & 0.2971    & 0.2654    & 0.2194 \\
layer.1.P  & 0.0648    & 0.1751    & 0.2908    & 0.2558    & 0.1966 \\
layer.0.P  & 0.0150    & 0.1262    & 0.2818    & 0.2466    & 0.1674 \\ \bottomrule
\end{tabular}
\caption{
The Pearson correlation ($r$) between BERTScore (\texttt{bert-base-multilingual-cased})  and human ratings. 
}
\label{tab:bertscore.bbmc}
\end{table*}

\begin{table*}[]
\centering
\begin{tabular}{lccccc}
\toprule
                             & \multicolumn{2}{c}{Zh-En} & \multicolumn{2}{c}{En-Zh} & \multirow{2}{*}{Ave.} \\
                             & 80\%          & 50\%          & 80\%          & 50\%          &                       \\\midrule
layer.11.F  & 0.4560    & 0.5447  & 0.2586    & 0.2446  & 0.3760 \\
layer.10.F  & 0.4558    & 0.5337  & 0.2584    & 0.2546  & 0.3756 \\
layer.8.F   & 0.4477 & 0.5251   & 0.2663 & 0.2596   & 0.3747 \\
layer.9.F   & 0.4459 & 0.5234   & 0.2603 & 0.2575   & 0.3718 \\
layer.11.R  & 0.4550    & 0.5281  & 0.2582    & 0.2308  & 0.3680 \\
layer.10.R  & 0.4518    & 0.5199  & 0.2565    & 0.2404  & 0.3671 \\
layer.7.F   & 0.4289 & 0.4882   & 0.2753 & 0.2710   & 0.3659 \\
layer.8.R   & 0.4425 & 0.5127   & 0.2615 & 0.2458   & 0.3656 \\
layer.9.R   & 0.4401 & 0.5114   & 0.2574 & 0.2478   & 0.3642 \\
layer.12.F  & 0.4267    & 0.5275  & 0.2499    & 0.2485  & 0.3631 \\
layer.11.P  & 0.4358    & 0.5273  & 0.2428    & 0.2378  & 0.3609 \\
layer.10.P  & 0.4370    & 0.5123  & 0.2436    & 0.2479  & 0.3602 \\
layer.7.R   & 0.4326 & 0.4793   & 0.2687 & 0.2583   & 0.3597 \\
layer.12.R  & 0.4289    & 0.5171  & 0.2538    & 0.2384  & 0.3596 \\
layer.8.P   & 0.4315 & 0.4968   & 0.2520 & 0.2505   & 0.3577 \\
layer.6.F   & 0.4096 & 0.4654   & 0.2803 & 0.2735   & 0.3572 \\
layer.9.P   & 0.4295 & 0.4984   & 0.2460 & 0.2460   & 0.3550 \\
layer.6.R   & 0.4210 & 0.4612   & 0.2679 & 0.2617   & 0.3529 \\
layer.5.F   & 0.3836 & 0.4467   & 0.2826 & 0.2650   & 0.3445 \\
layer.12.P  & 0.3991    & 0.5033  & 0.2322    & 0.2388  & 0.3434 \\
layer.7.P   & 0.3939 & 0.4515   & 0.2618 & 0.2600   & 0.3418 \\
layer.5.R   & 0.4010 & 0.4456   & 0.2693 & 0.2505   & 0.3416 \\
layer.4.R   & 0.3907 & 0.4256   & 0.2684 & 0.2451   & 0.3324 \\
layer.4.F   & 0.3653 & 0.4134   & 0.2819 & 0.2652   & 0.3315 \\
layer.3.R   & 0.3937 & 0.4270   & 0.2671 & 0.2316   & 0.3298 \\
layer.3.F   & 0.3621 & 0.4036   & 0.2792 & 0.2519   & 0.3242 \\
layer.2.R   & 0.3805 & 0.4100   & 0.2640 & 0.2405   & 0.3238 \\
layer.6.P   & 0.3506 & 0.4129   & 0.2700 & 0.2608   & 0.3236 \\
layer.1.R   & 0.3688 & 0.4104   & 0.2677 & 0.2418   & 0.3222 \\
layer.2.F   & 0.3447 & 0.3805   & 0.2744 & 0.2581   & 0.3144 \\
layer.1.F   & 0.3269 & 0.3757   & 0.2755 & 0.2555   & 0.3084 \\
layer.5.P   & 0.3049 & 0.3806   & 0.2718 & 0.2557   & 0.3032 \\
layer.0.R   & 0.3494 & 0.3474   & 0.2712 & 0.2409   & 0.3022 \\
layer.0.F   & 0.3074 & 0.3268   & 0.2765 & 0.2550   & 0.2914 \\
layer.4.P   & 0.2608 & 0.3235   & 0.2690 & 0.2595   & 0.2782 \\
layer.3.P   & 0.2188 & 0.2807   & 0.2633 & 0.2479   & 0.2527 \\
layer.2.P   & 0.1858 & 0.2460   & 0.2587 & 0.2521   & 0.2357 \\
layer.1.P   & 0.1543 & 0.2246   & 0.2553 & 0.2448   & 0.2198 \\
layer.0.P   & 0.0833 & 0.1525   & 0.2499 & 0.2404   & 0.1815 \\ \bottomrule
\end{tabular}
\caption{
The Pearson correlation ($r$) between BERTScore (\texttt{bart-large-mnli})  and human ratings. 
}
\label{tab:bertscore.blm}
\end{table*}

\begin{table*}[]
\centering
\begin{tabular}{lccccc}
\toprule
                            & \multicolumn{2}{c}{Zh-En} & \multicolumn{2}{c}{En-Zh} & \multirow{2}{*}{Ave.} \\
                             & 80\%          & 50\%          & 80\%          & 50\%          &                       \\\midrule
layer.41.F & 0.4798    & 0.5500    & 0.2571    & 0.2838    & 0.3927 \\
layer.41.R & 0.4853    & 0.5269    & 0.2586    & 0.2954    & 0.3915 \\
layer.13.R & 0.4702    & 0.5249    & 0.2974    & 0.2685    & 0.3902 \\
layer.16.R & 0.4617    & 0.5218    & 0.2863    & 0.2879    & 0.3895 \\
layer.14.R & 0.4693    & 0.5212    & 0.2932    & 0.2736    & 0.3893 \\
layer.40.F & 0.4764    & 0.5429    & 0.2552    & 0.2812    & 0.3889 \\
layer.15.R & 0.4625    & 0.5200    & 0.2882    & 0.2843    & 0.3887 \\
layer.42.F & 0.4847    & 0.5435    & 0.2542    & 0.2724    & 0.3887 \\
layer.17.R & 0.4643    & 0.5193    & 0.2828    & 0.2879    & 0.3885 \\
layer.12.R & 0.4672    & 0.5246    & 0.2959    & 0.2626    & 0.3876 \\
layer.40.R & 0.4837    & 0.5175    & 0.2569    & 0.2917    & 0.3874 \\
layer.42.R & 0.4883    & 0.5221    & 0.2535    & 0.2816    & 0.3864 \\
layer.13.F & 0.4577    & 0.5400    & 0.3001    & 0.2461    & 0.3860 \\
layer.18.R & 0.4605    & 0.5176    & 0.2795    & 0.2835    & 0.3853 \\
layer.43.F & 0.4823    & 0.5338    & 0.2568    & 0.2660    & 0.3847 \\
layer.16.F & 0.4507    & 0.5389    & 0.2866    & 0.2626    & 0.3847 \\
layer.14.F & 0.4564    & 0.5375    & 0.2950    & 0.2482    & 0.3843 \\
layer.15.F & 0.4524    & 0.5391    & 0.2875    & 0.2565    & 0.3839 \\
layer.12.F & 0.4528    & 0.5388    & 0.3003    & 0.2397    & 0.3829 \\
layer.44.F & 0.4804    & 0.5253    & 0.2603    & 0.2654    & 0.3828 \\
layer.11.R & 0.4572    & 0.5168    & 0.2918    & 0.2652    & 0.3828 \\
layer.39.F & 0.4724    & 0.5353    & 0.2545    & 0.2659    & 0.3820 \\
layer.17.F & 0.4493    & 0.5344    & 0.2832    & 0.2596    & 0.3816 \\
layer.35.R & 0.4787    & 0.5039    & 0.2667    & 0.2768    & 0.3815 \\
layer.39.R & 0.4788    & 0.5068    & 0.2574    & 0.2820    & 0.3812 \\
layer.43.R & 0.4858    & 0.5129    & 0.2546    & 0.2713    & 0.3812 \\
layer.37.R & 0.4775    & 0.5084    & 0.2655    & 0.2720    & 0.3809 \\
layer.38.R & 0.4750    & 0.5038    & 0.2602    & 0.2816    & 0.3801 \\
layer.45.F & 0.4772    & 0.5186    & 0.2630    & 0.2597    & 0.3796 \\
layer.34.R & 0.4794    & 0.5012    & 0.2683    & 0.2696    & 0.3796 \\
layer.37.F & 0.4706    & 0.5352    & 0.2635    & 0.2483    & 0.3794 \\
layer.36.R & 0.4761    & 0.5066    & 0.2651    & 0.2688    & 0.3792 \\
layer.44.R & 0.4861    & 0.5042    & 0.2568    & 0.2691    & 0.3791 \\
layer.33.R & 0.4774    & 0.5007    & 0.2681    & 0.2700    & 0.3790 \\
layer.31.R & 0.4730    & 0.5027    & 0.2761    & 0.2634    & 0.3788 \\
layer.35.F & 0.4727    & 0.5347    & 0.2653    & 0.2420    & 0.3787 \\
layer.38.F & 0.4657    & 0.5294    & 0.2586    & 0.2597    & 0.3783 \\
layer.33.F & 0.4746    & 0.5351    & 0.2675    & 0.2344    & 0.3779 \\
layer.34.F & 0.4739    & 0.5343    & 0.2667    & 0.2359    & 0.3777 \\
layer.36.F & 0.4706    & 0.5369    & 0.2643    & 0.2387    & 0.3776 \\ \bottomrule
\end{tabular}
\caption{
The Pearson correlation ($r$) between BERTScore (\texttt{deberta-xlarge-mnli})  and human ratings. We only show the top 40 results due to space limitation, 
}
\label{tab:bertscore.dxm}
\end{table*}

\begin{table*}[]
\centering
\begin{tabular}{lccccc}
\toprule
                            & \multicolumn{2}{c}{Zh-En} & \multicolumn{2}{c}{En-Zh} & \multirow{2}{*}{Ave.} \\

                             & 80\%          & 50\%          & 80\%          & 50\%          &                       \\\midrule
layer.13.F & 0.4160    & 0.5210    & 0.2651    & 0.2557    & 0.3645 \\
layer.16.F & 0.4432    & 0.5334    & 0.2443    & 0.2336    & 0.3636 \\
layer.16.P & 0.4439    & 0.5352    & 0.2409    & 0.2301    & 0.3625 \\
layer.14.F & 0.4265    & 0.5210    & 0.2531    & 0.2452    & 0.3615 \\
layer.12.F & 0.4100    & 0.5161    & 0.2707    & 0.2490    & 0.3614 \\
layer.13.P & 0.4171    & 0.5184    & 0.2580    & 0.2517    & 0.3613 \\
layer.15.F & 0.4276    & 0.5240    & 0.2525    & 0.2379    & 0.3605 \\
layer.14.P & 0.4302    & 0.5234    & 0.2475    & 0.2373    & 0.3596 \\
layer.15.P & 0.4272    & 0.5196    & 0.2459    & 0.2337    & 0.3566 \\
layer.17.F & 0.4307    & 0.5192    & 0.2440    & 0.2312    & 0.3563 \\
layer.11.F & 0.4004    & 0.4983    & 0.2755    & 0.2481    & 0.3556 \\
layer.17.P & 0.4309    & 0.5207    & 0.2384    & 0.2286    & 0.3547 \\
layer.10.F & 0.3969    & 0.4864    & 0.2781    & 0.2551    & 0.3541 \\
layer.19.F & 0.4197    & 0.5140    & 0.2492    & 0.2323    & 0.3538 \\
layer.18.F & 0.4232    & 0.5163    & 0.2485    & 0.2271    & 0.3538 \\
layer.12.P & 0.3952    & 0.5056    & 0.2621    & 0.2424    & 0.3513 \\
layer.18.P & 0.4191    & 0.5136    & 0.2440    & 0.2241    & 0.3502 \\
layer.19.P & 0.4075    & 0.5089    & 0.2394    & 0.2293    & 0.3463 \\
layer.12.R & 0.4058    & 0.4961    & 0.2586    & 0.2221    & 0.3456 \\
layer.10.R & 0.4030    & 0.4677    & 0.2654    & 0.2334    & 0.3424 \\
layer.11.R & 0.4015    & 0.4778    & 0.2639    & 0.2243    & 0.3419 \\
layer.11.P & 0.3785    & 0.4848    & 0.2645    & 0.2393    & 0.3418 \\
layer.9.F  & 0.3803    & 0.4591    & 0.2764    & 0.2498    & 0.3414 \\
layer.13.R & 0.3929    & 0.4906    & 0.2544    & 0.2256    & 0.3409 \\
layer.20.F & 0.3928    & 0.4856    & 0.2490    & 0.2350    & 0.3406 \\
layer.9.R  & 0.3990    & 0.4594    & 0.2676    & 0.2319    & 0.3395 \\
layer.15.R & 0.4065    & 0.4963    & 0.2425    & 0.2101    & 0.3389 \\
layer.10.P & 0.3679    & 0.4706    & 0.2665    & 0.2449    & 0.3375 \\
layer.19.R & 0.4177    & 0.4880    & 0.2418    & 0.2022    & 0.3374 \\
layer.16.R & 0.4181    & 0.4925    & 0.2310    & 0.2040    & 0.3364 \\
layer.14.R & 0.3981    & 0.4806    & 0.2415    & 0.2200    & 0.3351 \\
layer.8.F  & 0.3668    & 0.4380    & 0.2779    & 0.2549    & 0.3344 \\
layer.20.R & 0.4059    & 0.4765    & 0.2464    & 0.2052    & 0.3335 \\
layer.18.R & 0.4117    & 0.4874    & 0.2351    & 0.1971    & 0.3328 \\
layer.17.R & 0.4105    & 0.4804    & 0.2334    & 0.2014    & 0.3314 \\
layer.8.R  & 0.3781    & 0.4309    & 0.2691    & 0.2397    & 0.3295 \\
layer.21.F & 0.3797    & 0.4704    & 0.2416    & 0.2186    & 0.3276 \\
layer.7.F  & 0.3521    & 0.4115    & 0.2796    & 0.2513    & 0.3236 \\
layer.20.P & 0.3609    & 0.4644    & 0.2357    & 0.2320    & 0.3232 \\
layer.7.R  & 0.3688    & 0.4162    & 0.2686    & 0.2381    & 0.3229 \\ \bottomrule
\end{tabular}
\caption{
The Pearson correlation ($r$) between BERTScore (\texttt{roberta-large-mnli})  and human ratings. We only show the top 40 results due to space limitation, 
}
\label{tab:bertscore.rlm}
\end{table*}

\section{Significance Testing}
\label{app:sign-test}

\begin{figure*}[]
    \centering
    \begin{subfigure}[c]{0.45\textwidth}
        \centering
        \includegraphics[width=\textwidth]{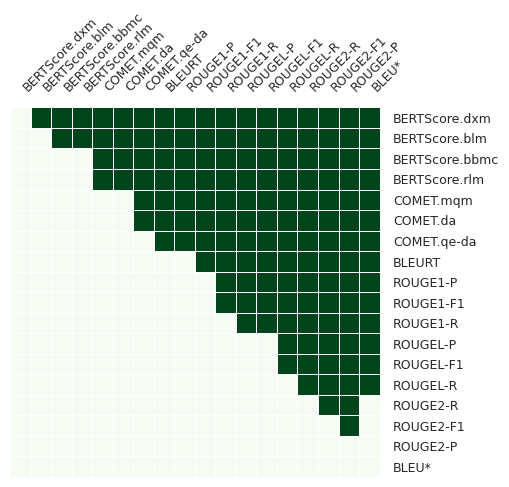}
        \caption{80\%-length, Zh-En}
        \label{fig:app-sub3}
    \end{subfigure}
    \hfill
    \begin{subfigure}[c]{0.45\textwidth}
        \centering
        \includegraphics[width=\textwidth]{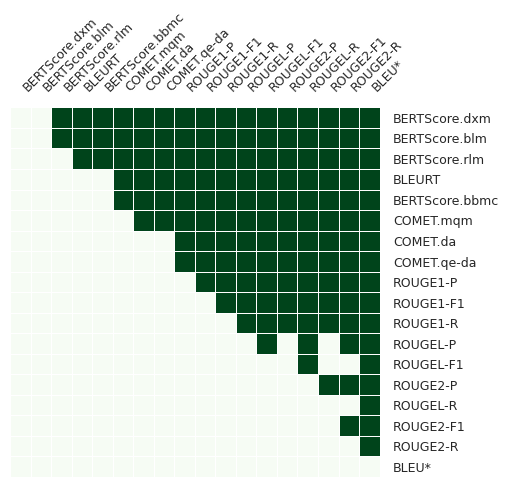}
        \caption{50\%-length, Zh-En}
        \label{fig:app-sub4}
    \end{subfigure}
    \caption{System-level significance tests. Green cells indicate a significant win for the row metric over the column metric, with 95\% confidence intervals of a difference in correlations not containing zero.}
    \label{fig:app-system-level}
\end{figure*}

\begin{figure*}[]
    \centering
    \begin{subfigure}[c]{0.45\textwidth}
        \centering
        \includegraphics[width=\textwidth]{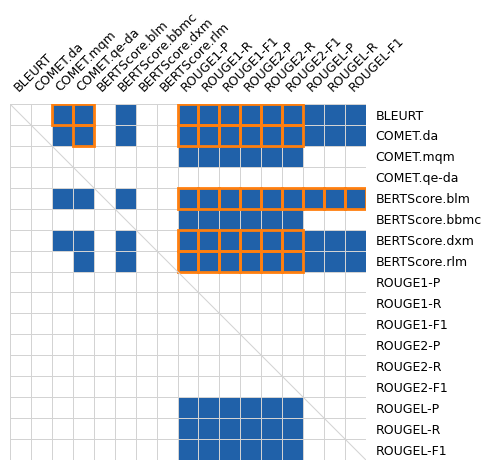}
        \caption{80\%-length, Zh-En}
        \label{fig:app-seg-sub3}
    \end{subfigure}
    \hfill
    \begin{subfigure}[c]{0.45\textwidth}
        \centering
        \includegraphics[width=\textwidth]{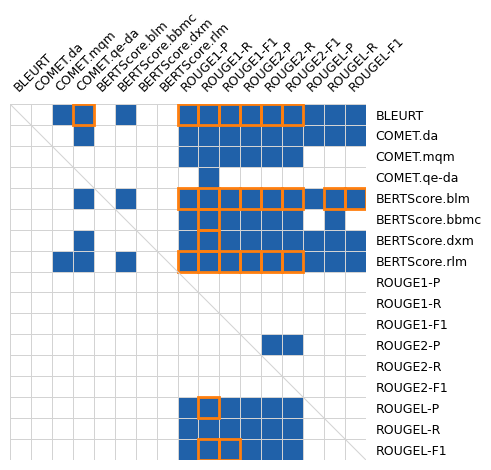}
        \caption{50\%-length, Zh-En}
        \label{fig:app-seg-sub4}
    \end{subfigure}
    \caption{Segment-level significance tests. 
Blue cells indicate the row metric has a higher correlation than the column metric ($p<0.05$).
The orange outline indicates the result remains significant after applying the Bonferroni correction.
}
    \label{fig:app-segment-level}
\end{figure*}

Figure~\ref{fig:app-system-level} shows the result of system-level significance testing on Zh-En. Figure~\ref{fig:app-segment-level} shows the result of segment-level significance testing on Zh-En.

\end{document}